\pdfoutput=1

\documentclass[11pt]{article}

\usepackage{acl}

\usepackage{times}
\usepackage{latexsym}

\usepackage[T1]{fontenc}

\usepackage[utf8]{inputenc}
\usepackage{colortbl}
\usepackage{amsmath,array,graphicx}
\usepackage{kantlipsum}
\usepackage{color}
\usepackage{adjustbox}
\usepackage{tabularray}

\usepackage{microtype}

\title{DTS-SQL: Decomposed Text-to-SQL with Small Large Language Models}

\author{Mohammadreza Pourreza \\
  University of Alberta  \\ 
  \texttt{pourreza@ualberta.ca} \\\And
  Davood Rafiei \\
  University of Alberta  \\
  \texttt{drafiei@uablerta.ca} \\}

\begin{document}
\maketitle
\begin{abstract}
Leading models for the text-to-SQL task heavily rely on proprietary Large Language Models (LLMs), posing concerns over data privacy. Closing the performance gap between small open-source models and large proprietary models is crucial to mitigate this reliance. To this end, we introduce a novel two-stage fine-tuning approach that decomposes the task into two simpler tasks. Through comprehensive evaluation on two large cross-domain datasets and two small LLMs, we show that this approach improves execution accuracy by 3 to 7 percent, effectively aligning the performance of open-source models with their proprietary counterparts.
\end{abstract}

\section{Introduction}
Natural language interfaces for databases allow users to derive insights from structured databases using natural language instead of complex SQL queries. Leading open-source methods \citep{pourreza2023din, gao2023text} for this task heavily depend on proprietary Large language models (LLMs) like GPT-4 and GPT-3.5-turbo, which have demonstrated superior performance in Text-to-SQL benchamrks \citep{yu2018spider, li2023can, gan-etal-2021-towards}. However, this reliance on large proprietary models has privacy and cost implications. For instance, many large enterprises cannot share their customer data  with the model-providing companies due to privacy considerations. Additionally, cost is a factor, especially for small businesses, in adopting these models.

Recent attempts to utilize open-source LLMs \citep{gao2023text} and fine-tune them using question-SQL query pairs have fallen short of the zero-shot performance of GPT-3.5-turbo. Table \ref{tab:1} presents a performance comparison of the fine-tuned open-source LLMs on the Spider development set, contrasting with methods that employ GPT-4's prompting techniques. This paper aims to address this disparity by introducing a novel two-step decomposed fine-tuning method, employing two smaller LLMs, each with a paramater size of 7 billion. This approach achieves a performance comparable to methods that are using GPT-4 with few-shot learning and well-designed prompts. 

\begin{table}
\centering
\small
\begin{tabular}{ccc}
\rowcolor[rgb]{0.502,0.502,0.502} \textbf{Model} & \textbf{EX} & \textbf{EM} \\
\multicolumn{3}{c}{{\cellcolor[rgb]{0.753,0.753,0.753}} Fine-tuning methods} \\
Llama2 7B \\ \citep{gao2023text} & 66.7 & 63.9 \\
Llama2 13B \\ \citep{gao2023text} & 67.0 & 62.7 \\
\multicolumn{3}{c}{{\cellcolor[rgb]{0.753,0.753,0.753}}Prompting methods} \\
DAIL-SQL + GPT4 \\ \citep{gao2023text} & 84.4 & 74.4 \\
DIN-SQL + GPT4 \\ \citep{pourreza2023din} & 74.2 & 60.1 \\
\end{tabular}
\caption{Perfromance comparison of the prompting methods and fineutning method on Spider validation dataset}
\label{tab:1}
\end{table}

We evaluate the performance of our proposed method using two Text-to-SQL benchmarks: Spider \citep{yu2018spider} and Spider-SYN \citep{gan-etal-2021-towards} and two 7B LLMs: DeepSeek \citet{deepseek-llm} and Mistral \citet{jiang2023mistral}. Our approach demonstrates a performance improvement of approximately 3 to 7 percent in execution accuracy compared to the conventional single-step fine-tuning method employed in previous studies \citep{gao2023text}. This consistent performance gain across both datasets highlights the generalizability of our method. Moreover, our fine-tuning strategy, utilizing a 7 billion parameter LLM, surpasses all previous open-source methods on the Spider development set and achieves comparable results to the state-of-the-art open-source methods using GPT-4 \citep{pourreza2023din, gao2023text} on the Spider test set. We have provided all the necessary code to replicate the results, along with the models' predicted SQL queries, in our GitHub repository \footnote{\url{https://anonymous.4open.science/r/DTS-SQL-2A42}}.

\section{Methodology}
\label{Methodology}

A notable development in LLMs is their post-pretraining refinement, which enhances their alignment with preferred behaviors, as documented by \citet{mishra2021cross, victor2022multitask, thoppilan2022lamda}. Common methods of alignment include Supervised Fine-Tuning (SFT) using human demonstrations, as reported by \citet{ouyang2022training, tunstall2023zephyr} and Reinforcement Learning from Human Feedback (RLHF), as detailed by \citet{christiano2017deep, ziegler2019fine, stiennon2020learning, bai2022training}.

The absence of extensive datasets containing either human or AI feedback \citep{lee2023rlaif} has led to a predominant focus on supervised fine-tuning in the text-to-SQL field. This approach necessitates a collection of specific instructions or prompts along with their corresponding outputs or responses. In the following section, we will delve into the established methods of supervised fine-tuning for LLMs within the Text-to-SQL context. Subsequently, we introduce our novel two-step fine-tuning approach, designed to enhance the performance of models in the Text-to-SQL domain.

\subsection{Supervised fine-tuning for Text-to-SQL}
\label{Supervised_finetuning}
In this section, we explore the supervised fine-tuning process for Text-to-SQL tasks, as practiced in the open-source community \citep{gao2023text}. Given a set of databases \( D_i \) comprising pairs of questions \( q_i \) and corresponding SQL queries \( s_i \), the goal is to fine-tune a large language model \( M \) using a set of training data \( T = \{(q_i, s_i, D_i)\} \), where \( q_i \) and \( s_i \) represent the natural language question and its associated SQL query on database \( D_i \). The objective of supervised fine-tuning is to minimize the empirical loss defined as: 

\begin{equation}
    \min_{\sigma, M^*} \frac{1}{|T|} \sum_{i=1}^{|T|} \mathcal{L}(M^*(\sigma_f(q_i, D_i, s_i)),
\end{equation}

where \( \mathcal{L} \) is the loss function used to measure the difference between the SQL queries generated by the model and the actual, correct (ground truth) queries. The function \( \sigma_f \) determines the formatting of the question, the database schema, and the SQL queries. A key challenge during inference is that we do not know in advance among all of the tables inside the database which tables are relevant to a given question for generating accurate SQL queries. Therefore, a common approach in fine-tuning involves including the schema of all tables within the prompts together with the question and SQL pairs. This method serves a dual purpose: teaching the model to generate the correct SQL query and to identify the relevant tables from among all the provided tables. This approach of training for two objectives simultaneously complicates the SQL generation task for LLMs, particularly for smaller models with only a few billion parameters. Each task – generating SQL queries and correctly linking to the relevant schema – demands its own reasoning process. A significant proportion of errors in large language models can be attributed to incorrect schema linking, highlighting this as a major challenge in the field \citep{pourreza2023din, dong2023c3}.  

\subsection{Decomposed Supervised Fine-tuning}

We propose a two-stage fine-tuning process, which separates schema linking and SQL generation, aiming to enhance the performance of NL-to-SQL systems.

\subsubsection{Schema-linking Fine-tuning}
Schema linking involves identifying the pertinent columns and tables in a database in response to natural language queries. It has been demonstrated to enhance cross-domain generalizability and facilitate the creation of intricate queries \citep{lei2020re}. In prior studies, schema linking has primarily been accomplished through in-context learning methods or implicitly during the fine-tuning process for SQL generation \citep{pourreza2023din, cao2021lgesql, guo2019towards, xu2021sead}.
In this work, we treat schema linking as a distinct task and explicitly fine-tune LLMs to identify relevant tables and columns when presented with a natural language query. Given the training dataset \( T = \{(q_i, s_i, D_i)\} \), we extract all of the columns and tables used in the SQL queries and create a new dataset of \( T = \{(q_i, T_i, C_i, D_i)\} \) where \( T_i \) and \( C_i \) represent lists of tables and columns used in the SQL query \( s_i \). The primary objective during supervised fine-tuning for schema linking is to minimize the empirical loss, as defined by the following equation:

\begin{equation}
    \min_{\sigma, M^*} \frac{1}{|T|} \sum_{i=1}^{|T|} \mathcal{L}(M^*(\sigma_s(q_i, T_i, C_i, D_i)),
\end{equation}

Here, \( \mathcal{L} \) represents the loss related to the model's next token prediction, comparing the predicted column and table names with the actual ground truth names.

\subsubsection{SQL Generation Fine-tuning}

After identifying the appropriate tables for SQL generation, the next step is to utilize a model that constructs the SQL query based on the question and the schema of the correct tables. Since we have already identified the potentially correct tables using the schema-linking module, there is no need to include all tables in the input for the SQL generation model. In contrast to previous approaches for fine-tuning LLMs, we extract the relevant tables from the training dataset \( T = \{(q_i, s_i, D_i)\} \) corresponding to the ground truth SQL queries. We then fine-tune the LLM while minimizing the following loss function:

\begin{equation}
    \min_{\sigma, M^*} \frac{1}{|T|} \sum_{i=1}^{|T|} \mathcal{L}(M^*(\sigma_g(q_i, T_i, s_i)),
\end{equation}

The loss function is same as the loss function defined in Section \ref{Supervised_finetuning}. This decomposition of the Text-to-SQL training process allows LLMs to be trained with a singular objective. By segregating the schema-linking and SQL query generation tasks, we improve the training process, enabling more focused and effective fine-tuning.

\section{Experiments}

\subsection{Models}

Our methodology's performance was assessed using two recent LLMs from distinct architectures. These models are Mistral 7B \citep{jiang2023mistral} and DeepSeek 7B \citep{deepseek-llm}. The DeepSeek model, sharing similar architecture with the LLama model family \citep{touvron2023llama}, has been pretrained on an extensive dataset comprising 2 trillion tokens and supports a sequence length of 4096. Mistral 7B, although not specifically pretrained for code generation, surpasses many counterparts in its scale category \citep{jiang2023mistral}.

\subsection{Hyperparameters}

The two LLMs were trained on Nvidia Tesla A100 GPUs, employing a batch sizes of 64 and 32 with a learning rate of 1*e-5 and 5*e-5 respectively. To enhance the training efficiency, we incorporated Flash Attention techniques as detailed in \citep{dao2022flashattention, dao2023flashattention}.

\subsection{Datasets}

We conducted our evaluation using cross-domain, challenging Text-to-SQL datasets. Spider, was introduced by \citet{yu2018spider} and includes 200 database schemas. Of these, 160 schemas are allocated for training and development, while the remaining 40 are set aside for testing purposes. Our second dataset was Spider-Syn \citep{gan-etal-2021-towards}, which modifies the original Spider dataset by replacing schema-related words with synonyms and removing explicit mentions that link natural language queries (NLQs) to the database schema.

\subsection{Metrics}

In our evaluation of text-to-SQL models, we utilized exact set match accuracy and execution accuracy. The former involves comparing the components of SQL queries, such as select, where, having, group by, and order by clauses, focusing on the matching of columns and predicates without considering the order. The latter determines equivalence between a model-generated query and a reference query if they produce identical results across various database instances.
\subsection{Results}
\label{results}

\subsubsection{Spider test set}

As depicted in Table \ref{tab:2}, our method employing DeepSeek 7B, when tested on the Spider test dataset, achieves results comparable to state-of-the-art open-source methods in terms of execution accuracy and exact set match accuracy.

\begin{table}
\centering
\small
\begin{tabular}{ccc}
\rowcolor[rgb]{0.502,0.502,0.502} \textbf{Model} & \textbf{EX} & \textbf{EM} \\ 
\cline{1-3}
\textcolor[rgb]{0.2,0.2,0.2}{DAIL-SQL + GPT-4}\\ {\citep{gao2023text}} & 86.6 & - \\
\textcolor[rgb]{0.2,0.2,0.2}{DIN-SQL + GPT-4} \\ {\citep{pourreza2023din}} & 85.3 & 60 \\
DTS-SQL + DeepSeek 7B \\ Ours & 84.4 & 73.7 \\
\textcolor[rgb]{0.2,0.2,0.2}{C3 + ChatGPT + Zero-Shot} \\ {\citep{dong2023c3}} & 82.3 & - \\
\textcolor[rgb]{0.2,0.2,0.2}{RESDSQL-3B + NatSQL} \\ {\citep{li2023resdsql}} & 79.9 & 72 \\
\textcolor[rgb]{0.2,0.2,0.2}{DIN-SQL + CodeX} \\ {\citep{pourreza2023din}} & 78.2 & 57 \\
DTS-SQL + Mistral \\ Ours & 77.1 & 69.3 \\
\textcolor[rgb]{0.2,0.2,0.2}{Graphix-3B + PICARD} \\  {\citep{li2023graphix}} & - & 74
\end{tabular}
\caption{The comparison of different methods on test set of Spider.}
\label{tab:2}
\end{table}

\subsubsection{Spider dev set}

In Table \ref{tab:3}, we showcase the results of our two-stage fine-tuning method on the dev set of Spider. The performance is compared against two distinct scenarios: firstly, a one-stage scenario where the model is fine-tuned on all tables without employing our two-stage approach, and secondly, a perfect schema linking scenario where we provide the ground truth tables to our fine-tuned SQL generators. This latter scenario is denoted as the 'Upper Bound' in the table. Our two-stage model's performance is measured by initially using our fine-tuned schema linker model to identify potentially relevant tables, which are then provided as context to the SQL generator model. 

In Table \ref{tab:4}, we offer a detailed comparison between our method and various other baseline approaches. For the baselines, we selected diverse methods from different families of approaches that are using LLMs and are available as open source. Our two-stage decomposed approach with DeepSeek 7B attained state-of-the-art performance on the Spider development set, surpassing all previous methods that utilized prompting techniques and fine-tuning. Additionally, the results of our two-stage method on Spider-SYN dataset is provided in the appendix \ref{sec:appendix} section.

\begin{table}
\centering
\small
\begin{tabular}{cccc}
\rowcolor[rgb]{0.502,0.502,0.502} \textbf{Model} & \textbf{Tuning} & \textbf{EX} & \textbf{EM} \\ 
Mistral 7B & FT Tuning & 71.9 & 70.9 \\
Mistral 7B & DTS-SQL & 78.6 & 73.3 \\
Mistral 7B & Upper bound & 86.6 & 80.7 \\
DeepSeek 7B & FT Tuning & 82.1 & 69.0 \\
DeepSeek 7B & DTS-SQL & 85.5 & 79.1 \\
DeepSeek 7B & Upper bound & 90.3 & 84.2
\end{tabular}
\caption{Performance of the LLMs with different tuning methods on Spider dev set. FT stands for Full tables finetuning, Upper bound performance is the performance which we can achieve with a perfect schema linking.}
\label{tab:3}
\end{table}

\begin{table}
\centering
\small
\begin{tabular}{ccc}
\rowcolor[rgb]{0.502,0.502,0.502} \textbf{Model} & \textbf{EX} & \textbf{EM} \\
\multicolumn{3}{c}{{\cellcolor[rgb]{0.753,0.753,0.753}}Instruction tuning methods} \\
DTS-SQL + Mistral 7B \\ (our) & 78.6 & 73.3 \\
DTS-SQL + DeepSeek 7B \\ (our) & \textbf{85.5} & \textbf{79.1} \\
Llama2 7B \\ \citep{gao2023text} & 66.7 & 63.9 \\
Llama2 13B \\ \citep{gao2023text} & 67.0 & 62.7 \\
\multicolumn{3}{c}{{\cellcolor[rgb]{0.753,0.753,0.753}}Prompting methods} \\
DIN-SQL + GPT4 \\ \citep{pourreza2023din}  & 74.2 & 60.1 \\
DIN-SQL + CodeX \\ \citep{pourreza2023din} & 69.9 & 57.2 \\
DAIL-SQL + GPT4 \\ \citep{gao2023text} & 84.4 & 74.4 \\
C3 + GPT-3.5 \\ \citep{dong2023c3} & 81.8 & -
\end{tabular}
\caption{Performance of different methods with LLMs on the dev set of Spider.}
\label{tab:4}
\end{table}

\subsubsection{Schema-linking Performance}

As discussed in Section \ref{Methodology}, our approach employs two LLMs: one for schema linking and another for SQL query generation. The schema-linking model plays a pivotal role in our pipeline, as inaccuracies in table detection could hinder the SQL generator's ability to formulate the correct SQL queries. We fine-tuned two models, based on the Deepseek and Mistral models, for schema linking. Evaluation metrics, including exact set match, precision, and recall, were used to assess their performance. Detailed information about these models on two distinct datasets can be found in Table \ref{tab:6}.

\begin{table}
\centering
\small
\begin{tabular}{ccccc}
\rowcolor[rgb]{0.502,0.502,0.502} \textbf{Model} & \textbf{Dataset} & \textbf{EX} & \textbf{PR} & \textbf{RE} \\
\textcolor[rgb]{0.2,0.2,0.2}{DeepSeek} & Spider & 93.1 & 98.4 & 97.7 \\
Mistral & Spider & 91.1 & 97.5 & 97.8 \\
DeepSeek & Spider-SYN & 87.6 & 94.6 & 94.7 \\
Mistral & Spider-SYN & 85.3 & 91.2 & 90.5
\end{tabular}
\caption{Performance of the schema-linker model on Spider and Spider-SYN dev sets. PR stands for Precision, RE is recall, and EX is exact set match accuracy.}
\label{tab:6}
\end{table}

\section{Discussion}

While our two-step approach has achieved state-of-the-art results on the development set of Spider and demonstrated comparable performance to larger models like GPT-4 on the test set, there is still significant room for improvement, particularly for the schema-linking models. Currently, our schema-linking models achieve roughly 90\% exact set match accuracy. However, as noted in Table \ref{tab:3}, the substantial gap between the upper bound performance of the SQL generator and that of DTS-SQL calls for further research into the schema-linking. .

\section{Conclusion}
Before our research, small open-source models lagged behind large proprietary models in performance on the text-to-SQL task. Our two-stage fine-tuning approach breaks down the task into two simpler components, enabling small open-source models to rival larger ones. Subsequent efforts could focus on enhancing the performance of these stages and exploring improved methods for transferring the output of one stage to the next.

\section*{Limitations}
This paper has placed its primary emphasis on enhancing the performance of both small large language models for Text-to-SQL task. However, there remains scope for further investigation and comparison of various techniques for schema-linking. Exploring approaches like retrieval methods or in-context learning when applied in conjunction with larger models such as GPT-4 for the schema-linking task could yield valuable insights into identifying the most effective methodologies for schema-linking.

\section*{Ethics Statement}
In this paper, we place a strong emphasis on the significance of ethical considerations in every aspect of our research, from its inception to its presentation. We wholeheartedly commit to adhering to the ACL Ethics Policy and upholding ethical principles and guidelines throughout our research journey.

We have taken proactive measures to minimize any potential biases or discriminatory elements in our research design, data selection, and interpretation of results. Our dedication to transparency, precision, and fairness in reporting our findings is unwavering, and we have duly acknowledged and cited the work of others to give proper credit.

By incorporating this ethics statement, we aim to underscore our unwavering commitment to conducting research with integrity, respecting ethical principles, and contributing responsibly to the advancement of knowledge in our field.

\section*{Acknowledgements}

\bibliography{anthology,custom}
\bibliographystyle{acl_natbib}

\newpage
\appendix

\section{Appendix}
\label{sec:appendix}

\subsection{Spider-SYN dataset}

To assess the efficacy of our proposed method, we evaluated its performance on the development set of Spider-SYN. Although Spider-SYN possesses a distinct training set, we opted to test our fine-tuned models directly on its development set, without any additional tuning on the Spider-SYN training set. The same performance gain is observed on this dataset (see Table \ref{tab:5}) even though the model was not directly trai
ned in this dataset.

\begin{table}
\centering
\begin{tabular}{cccc}
\rowcolor[rgb]{0.502,0.502,0.502} \textbf{Model} & \textbf{Tuning} & \textbf{EX} & \textbf{EM} \\ 
Mistral 7B & FT Tuning & 67.0 & 63.9 \\
Mistral 7B & DTS-SQL & 71.1 & 64.6 \\
Mistral 7B & Upper bound & 81.9 & 74.5 \\
DeepSeek 7B & FT Tuning & 70.4 & 56.6 \\
DeepSeek 7B & DTS-SQL & 76.2 & 68.9 \\
DeepSeek 7B & Upper bound & 85.5 & 78.1
\end{tabular}
\caption{Performance of the LLMs with different tuning methods on Spider-SYN dev set. FT stands for Full tables finetuning, Upper bound performance is the performance which we can achieve with a perfect schema linking.}
\label{tab:5}
\end{table}

\subsection{Prompt}

In conducting all our experiments on both models, we adhered to a standardized prompt format to ensure consistency and facilitate reliable comparisons. The chosen prompt format is well-established as effective in the Text-to-SQL domain, as demonstrated in prior research by \citet{gao2023text}. In this format, we provided information about the foreign key constraints, primary keys, and column types. Furthermore, to guide the models in understanding how data is stored within the database, our prompt incorporated three sample rows, showcasing data entries. 

The specific prompt used for our experiments is as follows:

\begin{figure}[ht]
\centering
  \includegraphics[width=0.5\textwidth]{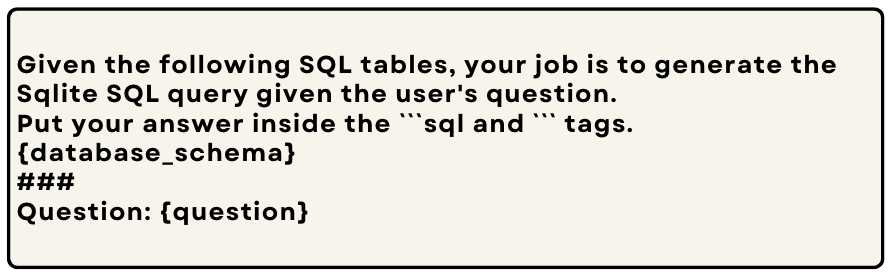}
  \caption{The prompt used for SQL generation. The database schema is where we put the tables representations.}
  \label{fig:1}
\end{figure}

\begin{figure}[ht]
\centering
  \includegraphics[width=0.5\textwidth]{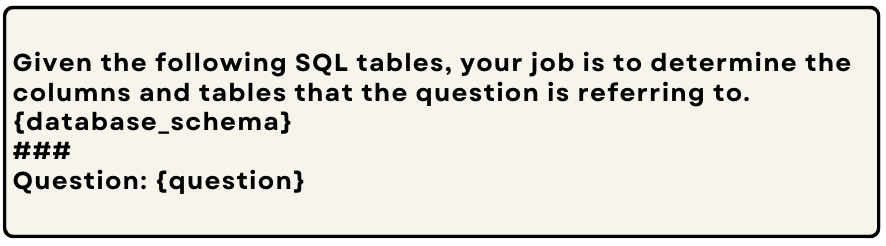}
  \caption{The prompt used for Schema linking. The database schema is where we put the tables representations.}
  \label{fig:2}
\end{figure}

\begin{figure}[ht]
\centering
  \includegraphics[width=0.5\textwidth]{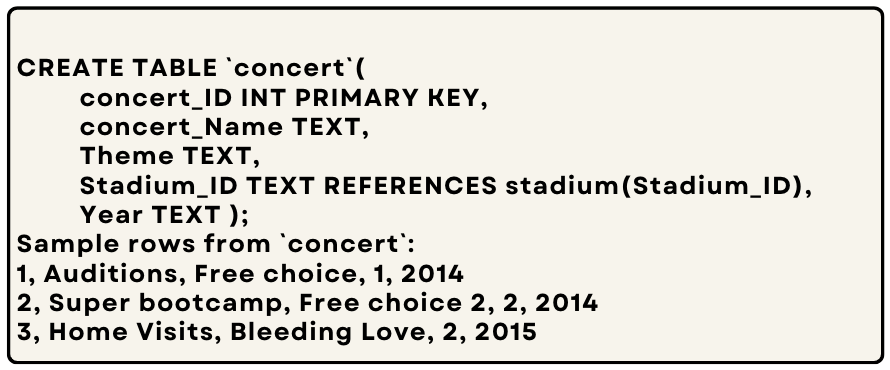}
  \caption{A sample table representation. All of the table in a database are represented as above and used in the prompts.}
  \label{fig:3}
\end{figure}

\end{document}